\documentclass[a4paper, 10pt, conference]{ieeeconf}      
\usepackage{FG2021}

\FGfinalcopy 

\IEEEoverridecommandlockouts                              
\usepackage{cite}
\usepackage{amsmath,amssymb,amsfonts}
\usepackage{algorithmic,algorithm}
\usepackage{graphicx}
\usepackage{textcomp}
\usepackage{xcolor}
\usepackage{multirow}
\usepackage{amsmath}
\usepackage{diagbox}
\usepackage{hyperref}

\def\FGPaperID{****} 

\title{\LARGE \bf
Modeling Dynamics of Facial Behavior for Mental Health Assessment
}

\author{\parbox{16cm}{\centering
    {\large Minh Tran*$^1$, Ellen Bradley*$^2$, Michelle Matvey$^2$, Joshua Woolley$^2$  and Mohammad Soleymani$^1$}\\
    {\normalsize
    $^1$ Computer Science Department, University of Southern California, Los Angeles, USA\\
    $^2$ University of California San Francisco, San Francisco, USA}}
    \thanks{*Both authors contributed equally to this research.}
}

\begin{document}

\ifFGfinal
\thispagestyle{empty}
\pagestyle{empty}
\else
\author{Anonymous FG2021 submission\\ Paper ID \FGPaperID \\}
\pagestyle{plain}
\fi
\maketitle

\newcommand{\etal}{\textit{et al}.}
\newcommand{\ie}{\textit{i}.\textit{e}.}
\newcommand{\eg}{\textit{e}.\textit{g}.}

\begin{abstract}

Facial action unit (FAU) intensities are popular descriptors for the analysis of facial behavior. However, FAUs are sparsely represented when only a few are activated at a time. In this study, we explore the possibility of representing the dynamics of facial expressions by adopting algorithms used for word representation in natural language processing. Specifically, we perform clustering on a large dataset of temporal facial expressions with 5.3M frames before applying the Global Vector representation (GloVe) algorithm to learn the embeddings of the facial clusters. We evaluate the usefulness of our learned representations on two downstream tasks: schizophrenia symptom estimation and depression severity regression. These experimental results show the potential effectiveness of our approach for improving the assessment of mental health symptoms over baseline models that use FAU intensities alone. 

\end{abstract}

\section{INTRODUCTION}

Representation learning is an increasingly popular field of research. In Natural Language Processing (NLP), pretrained language models such as BERT \cite{devlin2019bert} or word embeddings such as GloVe \cite{pennington2014glove} are normally used for initialization in downstream tasks such as sentiment analysis \cite{jianqiang2018deep}, name entity recognition \cite{chiu2016named} or question answering \cite{talmor2019commonsenseqa, yang2020bert}. 
However, learning facial representation, especially using FAUs, and investigating the effectiveness of the learned representation have been largely under-explored. Many studies use raw AU intensities as the features for their deep learning models in tasks such as deception detection \cite{rill2019high}, humor detection \cite{hasan-etal-2019-ur} or sentiment analysis \cite{yu2020ch}. Though this approach to facial representation provides important information about activated facial muscles, it is limited in that only a few FAUs are presented at a particular time (so most of the AU intensities are $0$). This is particularly problematic in research with clinical populations, where dataset sizes are often small due to the high costs associated with collecting data from patients.

Expressive deficits are core features of multiple neuropsychiatric disorders, including schizophrenia and depression \cite{tremeau2005facial}, that provide a high-yield window for assessing the degree of patients’ affective impairment. Unfortunately, expressive deficits are poorly understood and difficult to monitor, in large part because of a lack of effective measurement tools \cite{cohen2019using, insel2017digital}. Facial expressivity is typically quantified using interview-based clinical rating scales that yield ordinal values reflecting global impressions (\eg, “mild” or “moderate”) but cannot capture subtle changes in specific deficits that may offer clues about pathophysiology. In addition, administering these scales are costly, preventing the frequent measurement required to understand the dynamics of expressive behavior and to efficiently detect changes in a patient’s clinical status \cite{cohen2020ambulatory}.  These challenges impede our ability to (1) fully characterize the abnormalities that comprise expressive deficits to improve understanding of their functional and neural correlates; (2) develop and test new treatments for these deficits; and (3) effectively monitor patients’ symptoms over time and across settings. Thus, there is a critical need for quantitative, high-resolution, and efficient measurement techniques.

Behavioral measures such as FAUs that are too sparse to capture dynamics of expressions or too holistic to capture granular changes within facial regions are of limited utility \cite{alvino2007computerized}. We aim to fill this gap by developing a framework to learn an alternative representation of FAU intensities extracted from facial analysis toolkits, adapting the established GloVe embedding algorithm \cite{pennington2014glove} from the NLP domain. First, we perform \textit{k-means} clustering on a large dataset of more than $5.3M$ temporal FAU frames to create a ``vocabulary" of facial expressions, in which each cluster represent a type of facial expression. We then generate embeddings for these clusters to represent the ``meaning" of each cluster, similar to the way word embeddings can shed light on the semantics of words. This approach not only provides a dense representation of facial expressions but also helps ML models on low-resource datasets to learn faster and more accurately by utilizing the general knowledge about facial expressions from a larger dataset\footnote{Code is released at \url{https://github.com/mtran14/AUglove}}.

\section{RELATED WORK}
\subsection{Face representation}
Deep face embeddings are commonly used for face verification. Cao \etal~\cite{cao2010face} use learning-based descriptors and pose-adaptive matching to encode each face image with 26K descriptors before applying to PCA for compact face features. Chen \etal~\cite{chen2013blessing, chen2012bayesian} use a single-type Local Binary Pattern (LBP) descriptor to extract 100K facial features and use Joint Bayesian for face identification after PCA. The Facenet network (a deep CNN) is trained on a 260M images dataset using a triplet loss to convert facial images into 128-dimensional compact representations \cite{schroff2015facenet}. The model is reported to achieve up to 99.63\% face verification accuracy on the LFW dataset \cite{huang2008labeled}. Liu \etal~\cite{liu2017sphereface} approach face recognition using a deep hypersphere embedding approach. Their model learns discriminative face features with angular margin using the angular softmax loss for CNNs.

\subsection{Face clustering}
Zhou \etal~\cite{zhou2010unsupervised} present a novel temporal clustering algorithm, called Aligned Cluster Analysis, for clustering facial events from video of natural facial behaviors without recoursing to labeling schemes. De la Torre \etal~\cite{de2007temporal} propose a two-step approach for temporal segmentation of facial behavior. Their method first uses spectral graph techniques to cluster shape and appearance features before grouping the clusters into facial gestures. Vandal \etal present large-scale clustering of eyebrow raiser (AU02), eyebrow lowerer (AU04) and smile separately on 1.5 million videos of facial responses to online media contents \cite{vandal2015event}. They then use the clusters to find that smile events are more likely to occur during viral ads. Sen \etal~\cite{sen2018say} propose the Common Human Emotional Expression Set Encoder framework to identify combinations of facial action units relating to smiles that are well-represented by a small number of clusters. They find that clustering AU06+AU12 into 5 clusters can create clusters that are useful for deception detection. To the best of our knowledge, no prior work has attempted to perform large scale clustering on all action units.

\subsection{Word embedding models}
Word embedding models learn a mapping between words (or discrete values) into meaningful real-valued vectors. One line of word embedding research focuses on context-independent models. Collobert \etal~\cite{collobert2008unified} propose the lookup-table layer and demonstrate that word embeddings trained on sufficiently large data can carry semantic and syntactic meanings, which can improve the performance of downstream NLP tasks such as part-of-speech tagging, named entity recognition or language modeling. Mikolov \etal~\cite{mikolov2013efficient} use the \textit{skip-gram} model and utilize the idea that words appearing in similar locations have similar meanings to learn the word embeddings efficiently. Pennington \etal~\cite{pennington2014glove} present the Global Vectors (GloVe) model, which is able to derive semantic relationships between words from the co-occurrence matrix by maximizing the probability that a context word occurred given a center word. These context-independent models have been extended to other domains such as computational biology to represent protein sequences \cite{chen2019global} and DNA sequences \cite{min2017chromatin} or speech domain to represent audio segments \cite{chung2016audio, chung2018speech2vec}. Recently, another line of word embedding research has introduced successful context-dependent models such as BERT \cite{devlin2019bert} or ELMo \cite{peters2018deep}, which are capable of generating different embeddings for the same word appearing in different contexts. 

\section{METHOD}
Our general approach is to first perform clustering on a large dataset of temporal facial action unit (AU) intensities in order to obtain a diverse set of facial expressions as the cluster centers. Then, we use these cluster centers to train our embeddings.

\subsection{Dataset \& Feature Extraction}
In this study, we use the \texttt{VoxCeleb2} dataset to train our cluster embeddings \cite{chung2018voxceleb2}. \texttt{VoxCeleb2} contains more than 1M excerpts from over $6000$ celebrities collected from approximately $150K$ videos on YouTube; about $61\%$ of the speakers are men. We use OpenFace \cite{baltruvsaitis2016openface} to track FAUs in these videos \cite{baltrusaitis2018openface}. For each frame of an input video, OpenFace provides a set of 17 AU outputs as continuous values in the range of $[0,5]$, reflecting the intensity levels of the AUs. Frames with an OpenFace tracking confidence below $90\%$ are removed. We further downsample the OpenFace outputs to 5fps resulting in a dataset with more than $5.3M$ frames. 

\subsection{Clustering} \label{clusteringsection}
\begin{figure}[t]
  \centering
  \includegraphics[width=0.8\linewidth]{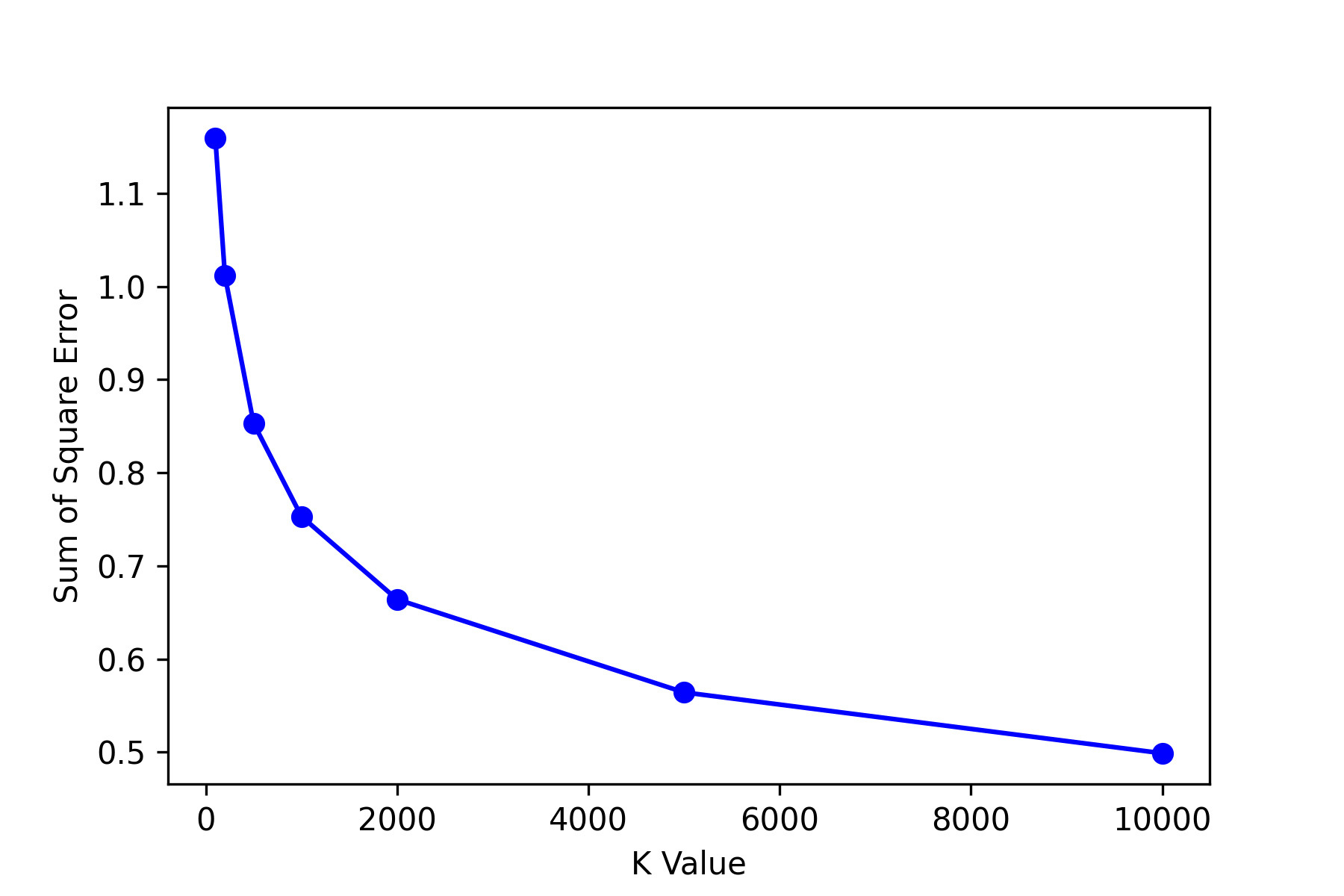}
  \vspace{-8pt}
  \caption{Elbow graph with different K values.}
  \label{figurek}
\end{figure}

\begin{figure*}[thpb]
  \centering
  \includegraphics[width=\linewidth]{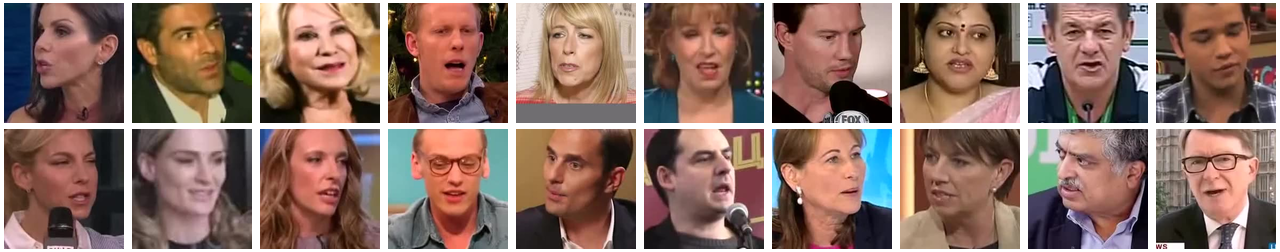}
  \vspace{-18pt}
  \caption{Images representing 20 randomly selected cluster centroids according to AU intensity (via OpenFace) are displayed. }
  \label{figure20centroids}
\end{figure*}

To identify a set of common facial expressions, we perform \textit{k-means} clustering on the $5.3M$ extracted facial expression frames from the above step.

Since the number of clusters (k) is a required parameter for the clustering algorithm, we perform clustering on a wide range of values $ k\in \{100, 200, 500, 1000, 2000, 5000, 10000\}$. 
In this study, we use the Elbow method to determine the optimal value for $k$. This method focuses on the rate of change of the average Euclidean distance between each data point and its closest cluster centers as the value of $k$ varies. Figure \ref{figurek} indicates the optimal value for $k$ to be either $1000$ or $2000$. We set $k=1000$ for further analysis. Due to the large size of data and high value of $k$, we use the \texttt{km-cuda} \footnote{https://github.com/src-d/kmcuda} library to speed up the clustering process. Figure \ref{figure20centroids} contains examples of real frames extracted from the \texttt{VoxCeleb2} dataset that are closest to 20 randomly chosen cluster centers.

\subsection{Embeddings training} \label{embeddingtrain}
We use the obtained cluster centers to transform the $5.3M$ frames of OpenFace AU outputs into their corresponding cluster labels. We add two special clusters, called \texttt{start} and \texttt{end}, to mark the starting and ending frames of each video. We also create a new cluster, called \texttt{unk}, to relabel a frame that falls into either one of the following cases 
\begin{enumerate}
    \item the cluster label of the frame appears less than $500$ times in the dataset
    \item the Euclidean distance between the frame and its corresponding cluster center is more than $1.75$
\end{enumerate}
We empirically determine these thresholds after observing the distribution of the number of occurrences of the clusters in the \texttt{VoxCeleb2} dataset as well as the distribution of the distance between frames and their closest cluster centers. The purpose of the \texttt{unk} cluster is to drop rare clusters that might be noisy for the embedding learning step or special facial expressions at a frame level that deviate from typical behaviors.

We then create the co-occurrence matrix with a window size of 10 (approximately 2 seconds before and after each frame) to identify the frequency at which one cluster label appears in the context of another, and use the GloVe model to learn the representation of each cluster label. In this study, we set the dimension of the GloVe embeddings to be in $\{25, 50, 100, 200\}$ and later explore the effects of embedding sizes on downstream task performances. 

\section{EXPERIMENT}
\subsection{Task \& Data} To show the usefulness of the proposed method, we apply the pretrained embeddings to two datasets: one dataset from people diagnosed with schizophrenia and another from people with depressive symptoms. In the schizophrenia dataset, we aim to estimate positive (psychotic) symptoms. In the depression dataset, we aim to detect participants' depression severity. 
\subsubsection{Schizophrenia dataset} The schizophrenia dataset was collected at the University of California, San Francisco. This dataset contains video recordings of 39 patients with schizophrenia who participated in a crossover clinical trial examining the effects of a single-dose oxytocin challenge on social behavior. Each patient was randomized to receive intranasal oxytocin (40 IU) on either testing day 1 or 2; on the other day they received a matched saline placebo. Following drug administration, patients completed an experimental task during which they viewed a set of brief video stimuli selected from Youtube as likely to evoke positive, negative or neutral emotions. Immediately after viewing each stimulus, patients were prompted to (1) describe the video, and (2) describe how the video made them feel. They had 30 seconds to respond to each prompt. We recorded videos of patients throughout the task. Patients were randomized to view one of two matched stimuli sets on each testing day. For this analysis, we extracted the segments from each video recording during which a participant was responding to the prompts. Next, we extracted the AU intensity levels for each of these segments at 5 fps with OpenFace. In total, this process produced 394 segments from patients on placebo and 437 segments on oxytocin. In addition to conducting the experimental task, we assessed the patients' positive symptom severity at baseline via standard clinical interview using the Positive and Negative Syndrome Scale (PANSS) \cite{kay1987positive}.

\subsubsection{Depression dataset} We use the dataset from the Audio/Visual Emotion Challenge and Workshop (AVEC 2019) \cite{ringeval2019avec}. In this study, we focus on the Detecting Depression with AI Sub-challenge (DDS). The dataset contains semi-clinical interviews designed to support the diagnosis of psychological distress conditions such as anxiety, depression, and post-traumatic stress disorder. Each interview (15-25 minutes) is self-rated with a depression severity score ranging from 0 to 24, which is obtained using the eight-item Patient Health Questionnaire (PHQ-8). The dataset is partitioned into a training set size of 163, validation set size of 56 and test set size of 56. We use the AU intensity levels for each of the interviews at 5 fps using the provided OpenFace files from the challenge.

\subsection{Models}

\begin{table*}[t]
\centering
\caption{Experiment results on the Schizophrenia Dataset.}
\vspace{-6pt}
\begin{tabular}{|c|c|c|c|c|c|c|c|c|c|c|}
\hline
\multirow{2}{*}{\textbf{Model}} & \multicolumn{2}{c|}{\textbf{Concept Disorg.}} & \multicolumn{2}{c|}{\textbf{Flow of Conv.}} & \multicolumn{2}{c|}{\textbf{Hallucinations}} & \multicolumn{2}{c|}{\textbf{Susp./persecutation}} & \multicolumn{2}{c|}{\textbf{Delusions}} \\ \cline{2-11} 
 & PCC $\uparrow$ & RMSE $\downarrow$ & PCC $\uparrow$ & RMSE $\downarrow$ & PCC $\uparrow$ & RMSE $\downarrow$ & PCC $\uparrow$ & RMSE $\downarrow$ & PCC $\uparrow$ & RMSE $\downarrow$\\ \hline
 Random & 0.127 & 2.256 & 0.130 & 2.192 & 0.133 & 2.387 & 0.131 & 2.069 & 0.132 & 2.256 \\
Continuous Dynamic Model & 0.418 & 1.415 & 0.473 & 1.27 & 0.516 & 1.313 & 0.303 & 1.271 & 0.45 & 1.312 \\ 
Static features+LR & 0.398 & 3.928 & 0.363 & 2.905 & 0.367 & 3.158 & 0.305 & 2.39 & 0.415 & 1.939 \\ 
LSTM+AU & 0.393 & 1.473 & 0.471 & 1.211 & 0.525 & 1.286 & 0.411 & 1.159 & 0.431 & 1.286 \\ \hline
LSTM+Cluster labels & \textbf{0.589} & 1.288 & 0.511 & 1.175 & 0.581 & 1.245 & 0.411 & 1.211 & 0.441 & 1.291 \\ 
LSTM+Cluster embeddings (d=25) & 0.504 & 1.329 & 0.571 & 1.158 & 0.666 & \textbf{1.139} & 0.519 & 1.096 & 0.51 & 1.242 \\ 
LSTM+Cluster embeddings (d=50) & 0.565 & 1.284 & 0.559 & 1.167 & 0.686 & 1.152 & 0.539 & 1.092 & 0.538 & 1.243 \\ 
LSTM+Cluster embeddings (d=100) & 0.559 & 1.301 & \textbf{0.613} & 1.117 & \textbf{0.688} & 1.155 & 0.564 & 1.067 & \textbf{0.587} & \textbf{1.208} \\ 
LSTM+Cluster embeddings (d=200) & 0.585 & \textbf{1.253} & 0.612 & \textbf{1.065} & 0.657 & 1.182 & \textbf{0.593} & \textbf{1.037} & 0.573 & 1.215 \\ \hline
\end{tabular}
\label{schz_regression_results}
\vspace{-16pt}
\end{table*}

\textbf{Continuous dynamic model.} This model, proposed by  Tron \etal~\cite{tron2015automated}, consists of a set static features obtained from facial Action Unit intensities that has been shown to be useful for schizophrenia symptom classification. For each facial AU in a given video, the signal is quantized into clusters that are then used to compute the activation ratio, activation level, activation length, change ratio and fast change ratio. This set of features is then used as the input for a Support Vector Regression (SVR). We use this model as a baseline for our schizophrenia symptom estimation experiment.

\textbf{Static Features + Linear Regression.} 
For each video recording, we compute the mean intensity, the mean of the first order derivative and the standard deviation for each AU. These features are then used as the input to a linear regression model to estimate a real-valued output. We use this model as a baseline for both the schizophrenia symptom estimation and depression severity regression experiments.

\textbf{LSTM}
We implement the Long-short Term Memory (LSTM) network to capture the temporal nature of the data. We change the inputs to the network to observe changes in performance with different types of AU representations. Specifically, inputs of the LSTM models can be: 

\begin{itemize}
    \item \textit{Resnet}: the face embeddings extracted from the ResNet-50 network \cite{he2016deep} that is pretrained on the AffWild dataset \cite{kollias2019deep}. This is only available for the depression dataset.
    \item \textit{AU}: the $17$ AU intensities extracted from OpenFace.
    \item \textit{Cluster labels}: sequences of cluster labels converted from the $17$ AU intensities using the cluster centers described in \ref{clusteringsection} (the embeddings of the cluster labels are initialized randomly).
    \item \textit{Cluster embeddings}: similar to the \textit{Cluster Labels} approach, but instead of initializing the embeddings of the cluster labels randomly we use the learned cluster embeddings described in Section \ref{embeddingtrain}.
\end{itemize}

\subsection{Experimental Setup}
For the schizophrenia experiment, we report the average scores from a 10-fold cross-validation on the stimuli segments. For each cross-validation iteration, we split a subset of the training set to use as an independent validation set. Due to space limitations, we show regression results for only $5$ of the PANSS items: conceptual disorganization, lack of spontaneity and flow of conversation, hallucinations, suspiciousness/persecution, and delusions. We select these items because they have the most balanced rating distributions in our dataset. 
We add a random baseline for both experiments, in which the predictions on the testing set are sampled based on the rating distribution of the training set. The results of the random baseline are averaged from 100 independent predictions.  
We report the Pearson Correlation Coefficient (PCC) and Root Mean-squared error (RMSE) for evaluation on both tasks. Following the baseline for the AVEC challenge \cite{ringeval2019avec}, we report the Concordance Correlation Coefficient (CCC) for the depression severity regression task. 

All LSTM models are bidirectional with a fixed hidden size of $32$ and a dropout rate of $0.4$. The final hidden state of the LSTM is passed into a fully connected layer with $64$ hidden units, followed by a single neuron to get the final regression output. We use the Adam optimizer with a learning rate of $1e^{-3}$ and a batch size of $8$. 
The hyper-parameters are selected according to the performance on the validation set. 
Embedding size is a hyper-parameter for \textit{LSTM+Cluster labels}. For the \textit{Continuous Dynamic Model}, we set $C$ and kernel types as hyper-parameters for SVR. To get stable results on the depression dataset, we train the LSTM models $50$ times and select the one with the highest CCC score on the validation set. 

\section{RESULT}
\begin{table}[t]
\begin{center}
\caption{Experiment results on the test set of the Depression Dataset.}
\begin{tabular}{|c|c|c|c|}
\hline
\textbf{Model} & \textbf{CCC} $\uparrow$ & \textbf{PCC} $\uparrow$ & \textbf{RMSE} $\downarrow$ \\ \hline 
Random & 0.086 & 0.093 & 8.973 \\ 
Static features + LR & 0.026 & 0.028 & 8.437 \\
LSTM + Resnet & 0.078 & 0.096 & 8.406 \\
LSTM + AU & 0.141 & 0.180 & 8.277 \\ \hline
LSTM + Cluster labels & 0.120 & 0.157 & \textbf{7.882} \\ 
LSTM + Cluster embeddings (d=25) &  \textbf{0.312} & \textbf{0.456} & 7.979 \\ 
LSTM + Cluster embeddings (d=50) & 0.258 & 0.349 & 8.174 \\ 
LSTM + Cluster embeddings (d=100) & 0.221 & 0.320 & 8.278 \\
LSTM + Cluster embeddings (d=200) & 0.237 & 0.334 & 8.164 \\ \hline
AU Baseline from AVEC \cite{ringeval2019avec} & 0.019 & - & 10.0 \\ 
Resnet Baseline from AVEC \cite{ringeval2019avec} & 0.120 & - & 8.01 \\ \hline
\end{tabular}
\label{table_depp}
\end{center}
\vspace{-16pt}
\end{table}
The regression results on the PANSS schizophrenia symptoms are provided in Table \ref{schz_regression_results}. The depression severity regression results are available in Table \ref{table_depp}. 
The models using the trained embeddings consistently outperform the models using the raw AU intensities as inputs on both tasks. In comparison with the baselines, our best-performing models achieve at least $0.15$ improvement in correlation scores on the estimation of schizophrenia symptoms task and $0.27$ better Pearson correlation on the depression severity regression task. Our models can also achieve up to $15\%$ lower RMSE in estimating schizophrenia symptoms. Further, we observe that for both tasks, the performance of \textit{LSTM + Cluster embeddings} is generally better than the performance of \textit{LSTM + Cluster labels}, which have the embeddings initialized randomly. This suggests the effectiveness of the pretrained embeddings in terms of capturing meaningful facial behavior descriptors.

The Schizophrenia dataset contains more videos with shorter duration while the Depression dataset contains less videos with longer duration. These differences might impact the performance of the embedding dimensions on the temporal models.

\section{CONCLUSION}
In summary, we propose a novel framework to represent FAU intensities. We first perform clustering on a large dataset of 5.3M frames to obtain a diverse set of facial expressions. Then, we use the GloVe embedding algorithm to learn the representations of these extracted facial expression clusters. We demonstrate the effectiveness of the learned embeddings on two downstream tasks, namely, schizophrenia symptom and depression severity estimation. Our methodological contribution may serve as an initial step in learning facial representations from sparse FAU intensity features, with important applications in neuropsychiatry research and clinical practice. Future work may investigate the possibility of applying successful context-dependent embedding models similar to BERT \cite{devlin2019bert} or ELMo \cite{peters2018deep} to learn FAU representations.

\section{ACKNOWLEDGMENTS}

Research was sponsored by the Army Research Office and was accomplished under Cooperative Agreement Number W911NF-20-2-0053. The views and conclusions contained in this document are those of the authors and should not be interpreted as representing the official policies, either expressed or implied, of the Army Research Office or the U.S. Government. The U.S. Government is authorized to reproduce and distribute reprints for Government purposes notwithstanding any copyright notation herein.



{\small
\bibliographystyle{ieee}
\bibliography{refs.bib}
}

\end{document}